\documentclass[runningheads]{llncs}
\usepackage[T1]{fontenc}
\usepackage{graphicx}
\usepackage{booktabs}
\usepackage{multirow}
\usepackage{ascmac}
\usepackage{amsmath}
\usepackage{bm}
\usepackage{amssymb}
\usepackage{comment}
\usepackage{algorithmic,algorithm}
\usepackage{indentfirst}
\usepackage{here}

\usepackage{color}

\usepackage{ulem}
\newcommand{\argmax}{\mathop{\mathrm{argmax}}\limits}

\newcommand{\defproposed}[1]{\textbf{{#1}}}
\newcommand{\defexisting}[1]{\textit{{#1}}}
\newcommand{\targetvar}[1]{\dashuline{#1}}
\newcommand{\targetdef}[1]{\uline{#1}}

\usepackage{hyperref}
\usepackage{color}

\begin{document}
\title{Data Augmentation Method Utilizing
Template Sentences for Variable Definition Extraction}
\titlerunning{Data Augmentation Method Utilizing Template Sentences}
%

\author{Kotaro Nagayama \and Shota Kato \and Manabu Kano}
\authorrunning{K. Nagayama et al.}

\institute{Graduate School of Informatics, Kyoto University \\
\email{nagayama.kotaro.63c@st.kyoto-u.ac.jp} \\
\email{\{shota, manabu\}@human.sys.i.kyoto-u.ac.jp}\\}

\maketitle

\begin{abstract}
The extraction of variable definitions from scientific and technical papers is essential for understanding these documents. However, the characteristics of variable definitions, such as the length and the words that make up the definition, differ among fields, which leads to differences in the performance of existing extraction methods across fields. Although preparing training data specific to each field can improve the performance of the methods, it is costly to create high-quality training data. To address this challenge, this study proposes a new method that generates new definition sentences from template sentences and variable-definition pairs in the training data. The proposed method has been tested on papers about chemical processes, and the results show that the model trained with the definition sentences generated by the proposed method achieved a higher accuracy of 89.6\%, surpassing existing models.

\end{abstract}

\section{Introduction}
The rate of increase in the number of papers worldwide has been significantly rising in recent years; it was around 5\% until 2017 and increased to 10.3\% in 2020 and 9.2\% in 2021~\cite{scirepo}.
Consequently, the time and effort required for literature review have been increasing year by year.
To alleviate this burden, several studies have proposed methods for automatically extracting, organizing, and utilizing information from a vast amount of literature using natural language processing (NLP) techniques~\cite{olivetti2020}. One of the most valuable information in scientific papers is mathematical formulas, which are used to express the relationships between variables.
Therefore, accurate extraction of variable definitions from scientific papers is essential for understanding the content of the papers.

Various methods for variable definition extraction have been proposed, including rule-based methods that extract noun phrases matching certain patterns~\cite{Pagel2014-nw}, machine learning-based methods using features such as part-of-speech tags and positions in sentences~\cite{Lin2019-ut,Schubotz2017-me}, and methods employing deep learning models~\cite{Stathopoulos2018-jl}.
Recently several methods have utilized pre-trained models such as Bidirectional Encoder Representations from Transformers (BERT)~\cite{bert} and achieved particularly high performance.
Kang et al.~\cite{kang} developed a method of extracting technical terms and their definitions simultaneously from texts using SciBERT~\cite{scibert} and attained an F1-score of 70.8\% in extracting definitions from 50 papers included in the ACL Anthology.
Lee et al.~\cite{Lee2022-re} tackled the task of matching variables to their definitions, SemEval 2022 Task 12: Symlink~\cite{symlink}. They used a simple rule-based tokenizer for symbols and SciBERT to perform named entity recognition (NER) and relation extraction (RE) of variables and their definitions sequentially~\cite{Lee2022-re}. Their method recorded the highest performance in the task. Popovic et al.~\cite{popovic} achieved the third highest performance in Symlink by employing SciBERT to simultaneously perform NER and RE.
Yamamoto et al.~\cite{yamamoto} aimed to develop a high-performance definition extraction method for chemical process-related datasets and proposed a method of predicting the position of a variable definition by inputting a sentence where the target variable is replaced by a special token into a BERT model.
They built a definition extraction model by fine-tuning DeBERTaV3$_\mathrm{LARGE}$~\cite{debertav3} in two steps: they first used a Symlink dataset~\cite{symlink} which has little relevance to chemical processes and then used a dataset related to chemical processes.
Their method achieved an accuracy of 85.5\% and an F1-score of 81.6\% in extracting variable definitions from papers on chemical processes.
Although these methods have achieved high performance, there are still challenges to be addressed in variable definition extraction.

Yamamoto et al.~\cite{yamamoto} achieved the highest performance among existing studies, but they used a smaller training dataset compared to datasets used in related research~\cite{kang,symlink,Lee2022-re,popovic}.
For example, Yamamoto et al. and Kang et al. used about 400 and 660 definition sentences for training, respectively~\cite{kang,yamamoto}.
Since these studies used a small amount of training data, increasing the amount of training data could potentially improve the performance of variable definition extraction.
However, the characteristics of variable definitions, such as the length and the words that make up the definition, vary by field, and thereby the difficulty level of variable definition extraction also differs among fields~\cite{popovic}.
Thus, increasing training data from other fields may not contribute to performance improvement. Additionally, it is very costly to create a large amount of new training data each time the field of application changes.

To solve these problems, this study proposes a method of augmenting the training data.
The proposed method prepares template sentences, such as ``[VAR\_1] is defined as [DEF\_1],'' and generates new definition sentences by assigning variables and their definitions in the training data to these template sentences.
Such a method of generating new data using existing data is generally called data augmentation.
Various methods have been proposed for natural language processing, but the conventional data augmentation methods focus on diversifying the meanings and expressions of sentences~\cite{li2022data}, which may not preserve the relationships between variables and their definitions.
In contrast, the proposed method preserves their expressions and the relationship between variables and their definitions, stabilizes the quality of the definition sentences, and diversifies only the sentence structure.
The performance of the model trained with the augmented training data using the proposed method is compared to that without data augmentation using a dataset created from 47 papers on chemical processes.

\section{Problem Setting}
In practice, the definitions of variables are sometimes scattered in figures and tables, and some of the definitions are not explicitly written in the text; thus, to understand all of the variables in a paper, one must refer to the paper and related documents to extract the definitions of the variables from the text, figures, and tables.
As the first step toward automating this process, this study tackles the task of extracting the definition of each variable from the main text of a paper if the definition exists and not extracting anything if it does not.
The positions of the variables are known in advance since the identification of variables is relatively easy~\cite{Lee2022-re}.
The target variables are those that meet the following three conditions: 1) the variable makes its first appearance in the paper, 2) it is surrounded by spaces or exists alone on the left side of an equation, and 3) when a corresponding definition exists in the text, it is in a form that can be directly extracted.
We set the first condition because the definition of a variable is often provided at its first mention, and the definitions of variables that appear multiple times in a paper are often abbreviated in subsequent mentions.
The second condition is set because variables that meet this condition are more likely to have their definitions written in the text.
We provide the third condition because some definitions are not directly extractable, which may affect the performance of the model.
For example, the definition of $C_\mathrm{A}$ and $C_\mathrm{B}$ in ``$C_\mathrm{A}$ and $C_\mathrm{B}$ are concentrations of A and B.'' are ``concentration of A'' and ``concentration of B,'' respectively.
Variables with such definitions cannot be directly extracted and could hinder the model's learning and evaluation; hence, such variables were set out of the scope.

\section{Proposed Method}
\begin{figure}[!t]
\centering
\includegraphics[width=\columnwidth]{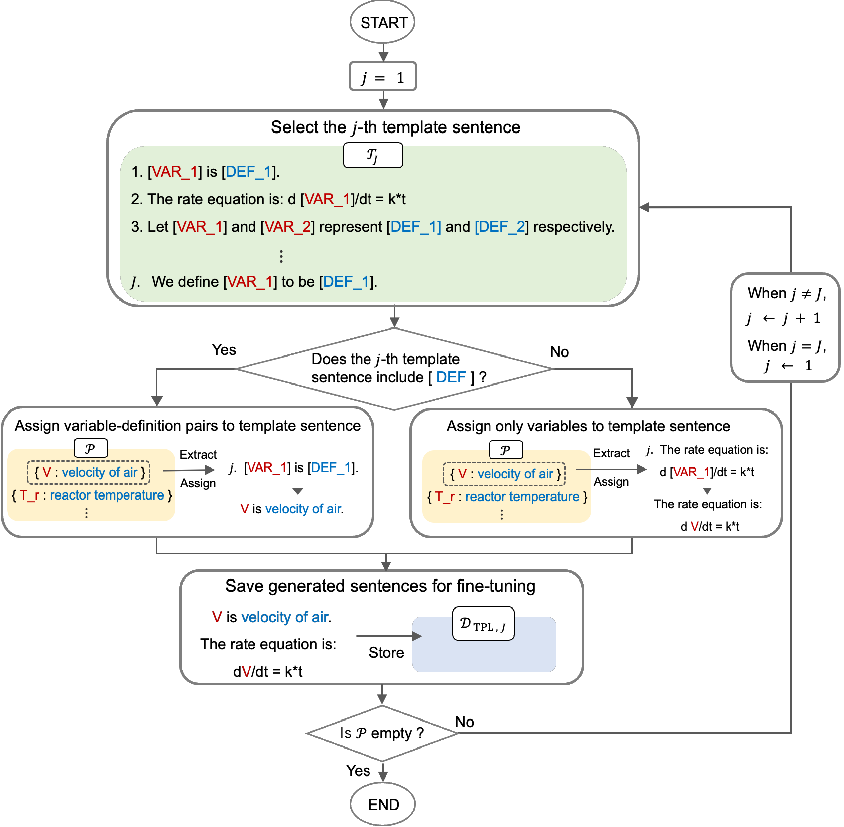}
\caption{Overview of the proposed method. $\mathcal{T}_J$ represents a set of $J$ types of template sentences, $\mathcal{P}$ is a list storing variable-definition pairs included in the training data, and $\mathcal{D}_{\mathrm{TPL},J}$ is a list for storing the generated sentences.}
\label{fig:method}
\end{figure}

\begin{figure}[t]
\centering
\includegraphics[width=.9\columnwidth]{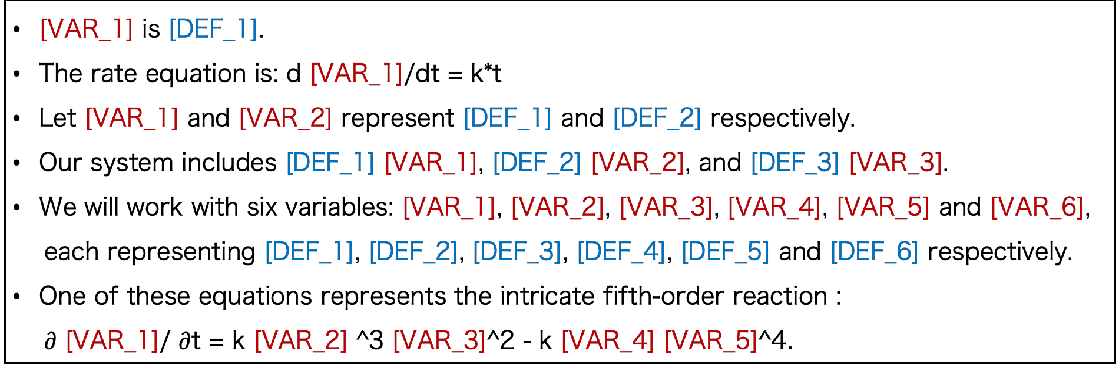}
\caption{Examples of template sentences, each containing pairs of variable and definition tokens or solely symbol tokens.
The number of variable tokens in each template sentence ranges from one to six.}
\label{fig:temp}
\end{figure}

The overview of the proposed method is illustrated in Fig.~\ref{fig:method}.
The proposed method prepares template sentences like those in Fig.~\ref{fig:temp} and generates new sentences from them and variable-definition pairs in training data.


The proposed method needs to prepare three pieces of information: a list $\mathcal{P}$ storing variable-definition pairs from the training data, a set $\mathcal{T}_J$ consisting of $J$ types of template sentences, and an empty list $\mathcal{D}_{\mathrm{TPL},J}$ for storing generated sentences.
The $j$-th template sentence in $\mathcal{T}_J$, TPL$_{j}$, includes $n_j$ pairs of variable and definition tokens (([VAR\_1], [DEF\_1]), $\ldots$, ([VAR\_j], [DEF\_j])), or $n_j$ variable tokens ([VAR\_1], $\ldots$, [VAR\_j]).
Template sentences are created using ChatGPT (gpt-3.5-turbo-1106).
The prompt is designed to include an explanation of the output format, specification of the number of variables in the generated sentence, and examples of definition sentences. 

The procedure for generating definition sentences is as follows:
\begin{enumerate}
\item Set $j=1$.
\item \label{enum:select_sentence} Select the $j$-th template sentence from $\mathcal{T}_J$, denoted as TPL$_{j}$.
\item Select $n_j$ variable-definition pairs from $\mathcal{P}$ and remove the pairs from $\mathcal{P}$.
\item Assign the selected $n_j$ variable-definition pairs to [VAR\_$i$] and [DEF\_$i$] ($i \in \{1, \ldots, n_j\}$) in TPL$_{j}$.
If TPL$_{j}$ does not include [DEF\_$i$], assign only the variables to [VAR\_$i$].
\item Add the generated sentence to $\mathcal{D}_{\mathrm{TPL},J}$.
\item Terminate data generation if $\mathcal{P}$ is empty. Otherwise, update $j$ to $j+1$ if $j\neq J$ or to 1 if $j=J$, and then return to step~\ref{enum:select_sentence}.
\end{enumerate}
The $\mathcal{D}_{\mathrm{TPL},J}$ created through the above steps contains the same number of variables as the variable-definition pairs included in the training data, with each being targeted for definition extraction once. Therefore, the number of variables contained in $\mathcal{D}_{\mathrm{TPL},J}$ does not depend on $J$.

\section{Experiment}
This section describes the variable definition extraction method, the datasets used for the experiments, the experimental setup, and the evaluation metrics.

\subsection{Variable Definition Extraction Method} \label{bert-de}
\begin{figure}[t]
  \includegraphics[width=\columnwidth]{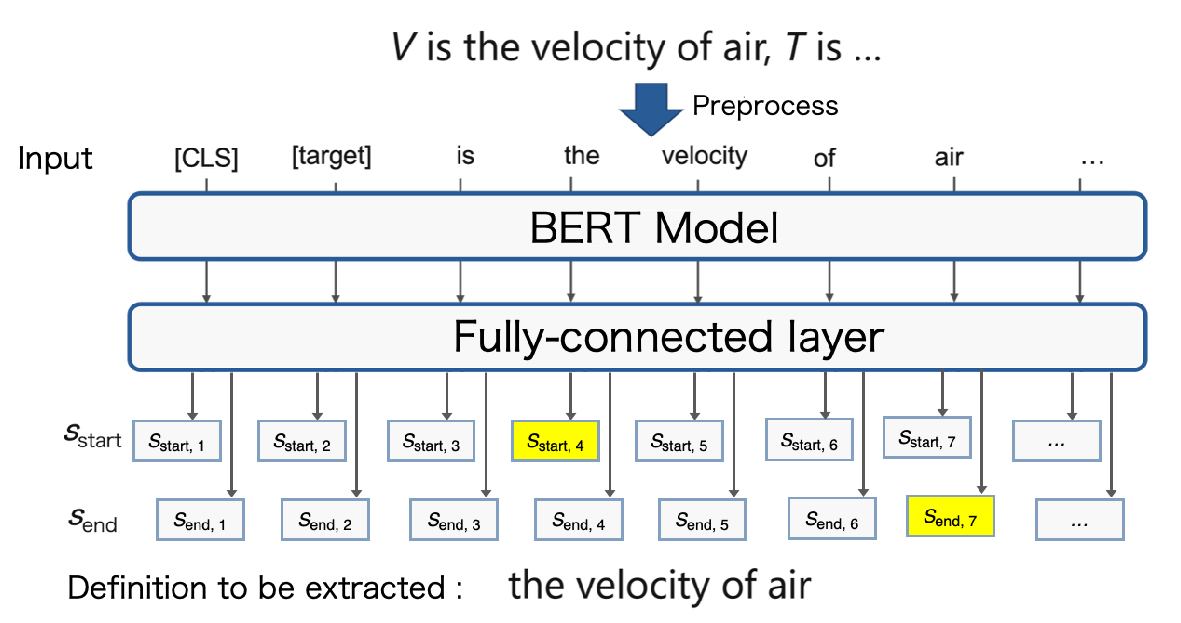}
  \caption{Schematic diagram of the variable definition extraction method by Yamamoto et al.~\cite{yamamoto}. The vectors $\bm{s}_{\mathrm{start}}$ and $\bm{s}_{\mathrm{end}}$ represent the probabilities of each token being the start and end positions of a definition, respectively. The tokens predicted as the start and end positions of a definition, corresponding to $s_{\mathrm{start},i}$ and $s_{\mathrm{end},j}$, are highlighted in yellow.}
  \label{fig:BERT}
\end{figure}

This section provides an overview of the variable definition extraction method developed by Yamamoto et al.~\cite{yamamoto}, which is used as a baseline in this study.
The schematic diagram of this method is shown in Fig.~\ref{fig:BERT}.
The method takes a sentence with a target variable for definition extraction and returns the corresponding definition if it exists or a [CLS] token if it does not exist.
Initially, the variable targeted for definition extraction within the given sentence is replaced with a special token [target].
Next, the sentence is split into tokens by a tokenizer.
The token sequence of length $d_{\mathrm{input}}$ is then provided to the BERT model and converted into context-dependent embedding vectors $\hat{\bm{e}_i}$ ($i \in \{1, 2, \ldots, d_{\mathrm{input}} \}$) by the BERT model.
These embedding vectors, $\hat{\bm{e}_i}$, are processed through a fully-connected layer and converted into $s_{\mathrm{start},i}$ and $s_{\mathrm{end},i}$, which are the probabilities that the corresponding tokens are the start and end positions of the definition respectively.
The method extracts as the definition the $k$-th to $l$-th tokens that satisfy the following conditions:
\begin{equation}
    k, l = \argmax_{k, l} s_{\mathrm{start},k} + s_{\mathrm{end},l}, \quad \mathrm{s.t.} \;  2 \leq k \leq l \; \mathrm{or}\;  k=l=1.
\end{equation}
If $s_{\mathrm{start},1} + s_{\mathrm{end},1}$ is the maximum, the method extracts the [CLS] token at the beginning of the input sentence, which indicates that no corresponding definition exists.

\subsection{Datasets} \label{sec:dataset}

We created a dataset, $\mathcal{D}_\mathrm{Process}$, which includes definitions for variables appearing in a total of 47 papers related to five processes: a crystallization process (CRYST), continuous stirred tank reactor (CSTR), biodiesel production process (BD), Czochralski process (CZ), and shell and tube heat exchanger (STHE).
The datasets of the processes are denoted as $\mathcal{D}_\mathrm{CRYST}$, $\mathcal{D}_\mathrm{CSTR}$, $\mathcal{D}_\mathrm{BD}$, $\mathcal{D}_\mathrm{CZ}$, and $\mathcal{D}_\mathrm{STHE}$, respectively.
The statistics of the datasets are summarized in Table~\ref{tab:process}.
$\mathcal{D}_\mathrm{Process}$ includes 1,214 variables, of which 67.5\% (820 variables) have definitions.

\begin{table}[t]
    \caption{The details of $\mathcal{D}_\mathrm{Process}$. \#Papers, \#Variables, and \#Variables with definitions represent the numbers of papers, variables, and variables with definitions.}
    \label{tab:process}
    \centering
    \begin{tabular}{lrrr}\toprule
    Dataset&\#Papers&\#Variables & \# Variables with definitions \\ \midrule
    $\mathcal{D}_\mathrm{CRYST}$& 11 & 281 & 200 \\
    $\mathcal{D}_\mathrm{CSTR}$& 10 & 169 & 123 \\
    $\mathcal{D}_\mathrm{BD}$& 10 & 186 & 125 \\
    $\mathcal{D}_\mathrm{CZ}$& 9 & 311 & 196 \\
    $\mathcal{D}_\mathrm{STHE}$& 7 & 267 & 176 \\ \midrule
    $\mathcal{D}_\mathrm{Process}$ & 47 & 1,214 & 820 \\ \bottomrule
    \end{tabular}
\end{table}

Similar to Yamamoto et al.~\cite{yamamoto}, we also use the dataset used in Symlink~\cite{symlink} denoted by $\mathcal{D}_\mathrm{Symlink}$.
$\mathcal{D}_\mathrm{Symlink}$ consists of 101 papers from five fields: informatics, biology, physics, mathematics, and economics, and contains 16,642 variables, of which 11,462 have definitions.

Each paper in $\mathcal{D}_\mathrm{Process}$ is classified as either part of training, validation, or test set.
In $\mathcal{D}_{\mathrm{STHE}}$, the number of papers designated for the test set is two, whereas for other datasets, this number is set to three.
For each dataset, one paper is assigned for validation and the remainder for training.
$\mathcal{D}_\mathrm{Symlink}$ and $\mathcal{D}_{\mathrm{TPL},J}$ are split into training and validation sets in a 3:1 ratio.
The training and validation datasets are used for fine-tuning models, while the test dataset serves for performance evaluation.

\subsubsection{}
To investigate the influence of the number of template sentences on definition extraction performance, we create three sets of template sentences: $\mathcal{T}_{20}$, $\mathcal{T}_{100}$, and $\mathcal{T}_{300}$, along with their corresponding datasets $\mathcal{D}_{\mathrm{TPL,20}}$, $\mathcal{D}_{\mathrm{TPL,100}}$, and $\mathcal{D}_{\mathrm{TPL,300}}$.
We initially construct 300 template sentences $\mathcal{T}_{300}$ by ensuring the proportion of definition tokens matches that in $\mathcal{D}_\mathrm{Process}$.
Subsequently, we randomly select 100 sentences from $\mathcal{T}_{300}$ to create $\mathcal{T}_{100}$, so that the distribution of the number of definition tokens remains consistent with $\mathcal{T}_{300}$.
Similarly, $\mathcal{T}_{20}$ is derived from $\mathcal{T}_{100}$.
Table~\ref{tab:tpl} summarizes their distributions of the number of definition tokens.
\begin{table}[t]
    \caption{Breakdown of the number of template sentences in $\mathcal{T}_J$}
    \label{tab:tpl}
    \centering
    \setlength{\tabcolsep}{8pt}
    \begin{tabular}{lrrrrrrr}
    \toprule
    \multirow{2}{*}{Dataset} & \multicolumn{7}{c}{Number of definition tokens} \\
    \cmidrule(lr){2-8}  &  0  & 1  & 2  & 3  & 4  & 5  & 6 \\ \midrule
    $\mathcal{T}_{20}$  & 8   & 2  & 2  & 2  & 2  & 2  & 2 \\
    $\mathcal{T}_{100}$ & 40  & 14 & 14 & 8  & 8  & 8  & 8  \\
    $\mathcal{T}_{300}$ & 120 & 42 & 42 & 24 & 24 & 24 & 24  \\
    \bottomrule
    \end{tabular}
\end{table}

\subsubsection{}
The proposed method can be utilized to extract definitions from papers on processes other than the five processes included in $\mathcal{D}_\mathrm{Process}$.
To verify the effectiveness of the proposed method when the target process is not contained in the training dataset, we make the dataset that does not include the dataset of process $X$, $\mathcal{D}_{\mathrm{Process}-X}$, by removing the dataset of the process $X$, $\mathcal{D}_X$, from $\mathcal{D}_\mathrm{Process}$. We train the model with $\mathcal{D}_{\mathrm{Process}-X}$ and then verify its definition extraction performance on $\mathcal{D}_X$.

\subsection{Experimental Setup}

We use DeBERTaV3$_\mathrm{LARGE}$~\cite{debertav3} as the base model, Adam~\cite{adam} as the optimizer, and NVIDIA A100 80GB for the GPU.
The batch size and learning rate are set to 8 and 1e-5, respectively.
Additionally, similar to the two-step fine-tuning performed by Yamamoto et al.~\cite{yamamoto}, we apply a three-step fine-tuning strategy using $\mathcal{D}_\mathrm{Symlink}$, $\mathcal{D}_{\mathrm{TPL},J}$, and
$\mathcal{D}_\mathrm{Process}$ in sequence.
Hereafter, the three-step fine-tuning using $\mathcal{D}_{\mathrm{TPL},J'}$, created from $J'$ types of template sentences, is referred to as ``Proposed$_{J=J'}$''.
We train the model for three epochs and apply the model with the minimum cross-entropy loss to the test data to evaluate the performance.

\subsection{Evaluation Method}
For each target variable, the variable definition extraction method yields a token sequence as a definition or the [CLS] token.
To evaluate the performance of the variable definition extraction method, we classify the output of the method into five categories: true positive (TP), false positive when the definition is extracted (FP$^{(1)}$), false positive when the [CLS] token is extracted (FP$^{(2)}$), false negative (FN), and true negative (TN).
These categories are organized in Table \ref{tab:index}.
\begin{table}[t]
    \caption{Classes of the outputs of the variable definition extraction method}
    \centering
    \begin{tabular}{lp{2em}lp{3em}l}
    \toprule
    & \multicolumn{4}{c}{Does the variable have a definition?} \\
    Result                && Yes        && No         \\ \midrule
    The extracted definition is correct.       && TP       && ---        \\
    The extracted definition is incorrect.     && FP$^{(1)}$ && FP$^{(2)}$ \\
    The [CLS] token is extracted.       && \multirow{2}{*}{FN}       && \multirow{2}{*}{TN}       \\
    (No definition is extracted) &&            &&            \\
    \bottomrule
    \end{tabular}
    \label{tab:index}
\end{table}
Using this classification, we define four metrics: Accuracy (Acc.), Precision (Pre.), Recall (Rec.), and the F1-score (F1). Acc. is defined as the proportion of variables for which the correct answer is obtained. Pre. represents the proportion of correctly extracted definitions out of all extracted definitions. Rec. denotes the proportion of correctly extracted definitions out of the variables with definitions. The F1 is calculated as the harmonic mean of $Pre.$ and $Rec.$ The equations for calculating these metrics are as follows:

\begin{equation}
  Acc. = \frac{TP + TN}{TP + FP^{(1)} + FP^{(2)} + FN + TN},
\end{equation}
\begin{equation}
  Pre. = \frac{TP}{TP + FP^{(1)} + FP^{(2)}},
\end{equation}
\begin{equation}
  Rec. = \frac{TP}{TP + FP^{(1)} + FN},
\end{equation}
\begin{equation}
  F1 = \frac{2 \times Pre. \times Rec.}{Pre. + Rec.}.
\end{equation}
When performing variable definition extraction before other downstream tasks, both extracting the correct definition and correctly predicting the absence of a definition are important.
Therefore, the superiority of a method is judged by $Acc.$, and other metrics are used to grasp the characteristics of the method.

Since the performance can vary depending on the types of papers included in the training, validation, and test sets, we split $\mathcal{D}_\mathrm{Process}$ ten times according to Section~\ref{sec:dataset} and compare the average of their evaluation metrics.

\section{Results and Discussion}
\subsection{Results}
\begin{figure}[t]
    \centering
    \includegraphics[width=\columnwidth]{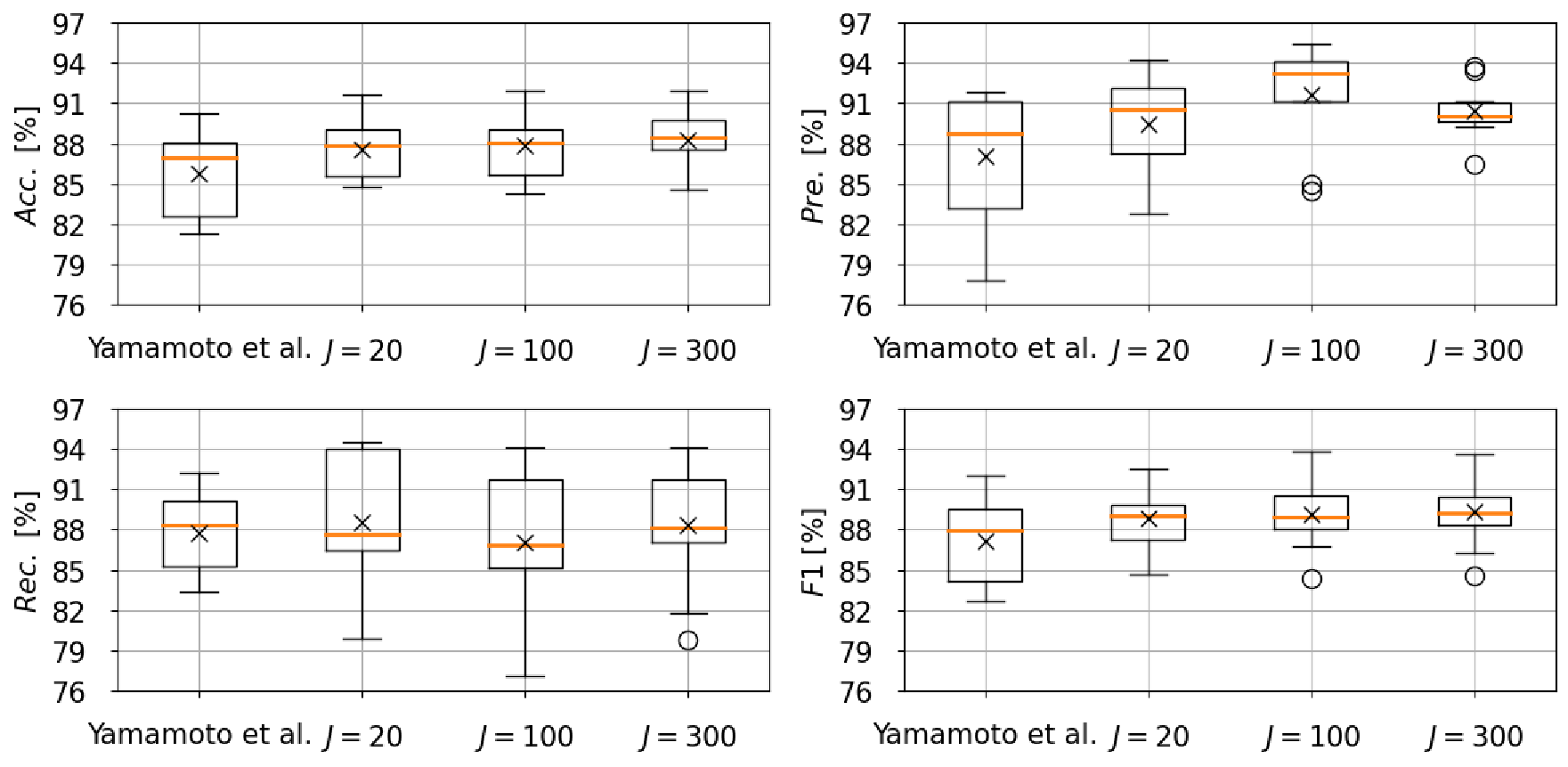}
    \caption{Distribution of each evaluation metric across ten experiments.  $J=J'$ in the figure corresponds to the distribution of the Proposed$_{J=J'}$.
    In each box, the orange horizontal line represents the median and the mark $\times$ represents the mean value.}
    \label{fig:hige}
\end{figure}

Table~\ref{table:result} shows the average evaluation metrics, and Fig.~\ref{fig:hige} shows the distribution of each evaluation metric across ten experiments.
In all evaluation metrics except for $Rec.$ when $J=100$, the proposed method outperformed the method of Yamamoto et al., demonstrating the effectiveness of the proposed method in improving variable definition extraction performance.
In $Acc.$, $Pre.$, and $F1$, the proposed method exhibited less variability compared to the method of Yamamoto et al., suggesting that the proposed method not only enhances variable definition extraction performance but also stabilizes the performance.
Although the variability in $Rec.$ was greater than that in the method of Yamamoto et al., the proposed method overall achieved higher performance.

\begin{table}[t]
    \caption{Experimental results of variable definition extraction.
    Each value is the average of ten experiments and is presented as a percentage.
    The bold number indicates the highest value in each column.}
    \label{table:result}
    \setlength{\tabcolsep}{12pt}
    \renewcommand{\arraystretch}{1.3}
    \centering
    \begin{tabular}{lrrrr} \toprule
    Method & $Acc.$ & $Rec.$ & $Pre.$ & $F1$  \\ \midrule
    Yamamoto et al.~\cite{yamamoto} & 85.8 & 87.4 & 87.1 & 87.1  \\
    Proposed$_{J=20}$& 87.6 & \textbf{88.5} & 89.4 & 88.8  \\
    Proposed$_{J=100}$& 87.8 & 87.1 & \textbf{91.6} & 89.1 \\
    Proposed$_{J=300}$ & \textbf{88.3} & 88.3 & 90.4 & \textbf{89.3}  \\ \bottomrule
    \end{tabular}
\end{table}

\subsection{Discussion}
\subsubsection{Impact of the Number of Template Sentences}

Both $Acc$. and $F1$ improved as the number of template sentences $J$ increased.
Among the cases where definition extraction failed by the Proposed$_{J=20}$  but succeeded by the Proposed$_{J=300}$, some definitions did not fit the patterns included in $\mathcal{T}_{300}$.
This suggests that increasing $J$ contributes not only to the performance improvement in definition extraction from the existing template patterns but also from patterns not covered by the templates.

\subsubsection{Performance on Unknown Processes}
Table~\ref{table:result_2} shows the variable definition extraction results of Proposed$_{J=300}$ and the method of Yamamoto et al. for unknown processes.
Across all five processes, regardless of whether the data of the target process was included in the training data, the proposed method outperformed those using the method of Yamamoto et al.
This demonstrates the effectiveness of the proposed method in extracting variable definitions even for processes not included in the training data.

\begin{table}[t]
  \caption{Variable definition extraction performance on unknown processes}
  \label{table:result_2}
  \centering
  \setlength{\tabcolsep}{4.5pt}
  \begin{tabular}{llrrrr|rrrr} \toprule
  \multirow{2}{*}{Train dataset} & \multirow{2}{*}{Test dataset} & \multicolumn{4}{c|}{Yamamoto et al.~\cite{yamamoto}} & \multicolumn{4}{c}{Proposed$_{J=300}$}\\
  \cmidrule{3-6} \cmidrule{7-10}
    & & $Acc.$& $Rec.$ &$Pre.$ & $F1$  & $Acc.$& $Rec.$ &$Pre.$ &$F1$ \\
    \midrule
    $\mathcal{D}_\mathrm{Process - CRYST}$ & \multirow{2}{*}{$\mathcal{D}_\mathrm{CRYST}$} & 78.0  & 74.8 & 81.7 & 78.0 & 82.0 & 79.7 & 85.1 & 82.2 \\
  $\mathcal{D}_\mathrm{Process}$ &  & 83.1  & 82.8 & 85.6 & 84.2 & 85.3 & 84.9 & 88.2 & 86.4 \\ \midrule
    $\mathcal{D}_\mathrm{Process - CSTR}$ & \multirow{2}{*}{$\mathcal{D}_\mathrm{CSTR}$} & 93.2  & 92.0 & 91.7 & 91.8 & 94.8 & 94.5 & 94.6 & 94.5 \\
    $\mathcal{D}_\mathrm{Process}$ &  &94.8  & 93.6 & 94.5 & 94.0 & 95.9 & 95.5 & 95.6 & 95.5  \\ \midrule
  $\mathcal{D}_\mathrm{Process - BD}$ & \multirow{2}{*}{$\mathcal{D}_\mathrm{BD}$} & 87.5  & 88.5 & 85.7 & 87.0 & 91.2 & 89.3 & 91.3 & 90.2 \\
  $\mathcal{D}_\mathrm{Process}$ &  & 90.9  & 88.7 & 91.8 & 90.2 & 91.6 & 91.7 & 92.0 & 91.7  \\ \midrule
  $\mathcal{D}_\mathrm{Process - CZ}$ & \multirow{2}{*}{$\mathcal{D}_\mathrm{CZ}$} & 82.8  & 91.1 & 79.5 & 84.9 & 86.4 & 89.6 & 86.5 & 87.9
  \\
  $\mathcal{D}_\mathrm{Process}$ &  & 86.2 & 89.6 & 86.3 & 87.8 & 89.7 & 89.4 & 90.9 & 90.1 \\\midrule
  $\mathcal{D}_\mathrm{Process - STHE}$ & \multirow{2}{*}{$\mathcal{D}_\mathrm{STHE}$} & 74.1  & 85.4 & 74.8 & 78.0 & 79.5 & 83.9 & 81.3 & 82.2 \\
  $\mathcal{D}_\mathrm{Process}$ &  & 79.6  & 84.4 & 82.3 & 82.6 & 83.7 & 83.6 & 88.1 & 85.4 \\ \bottomrule
  \end{tabular}
\end{table}

\begin{figure}[t]
  \includegraphics[width=\columnwidth]{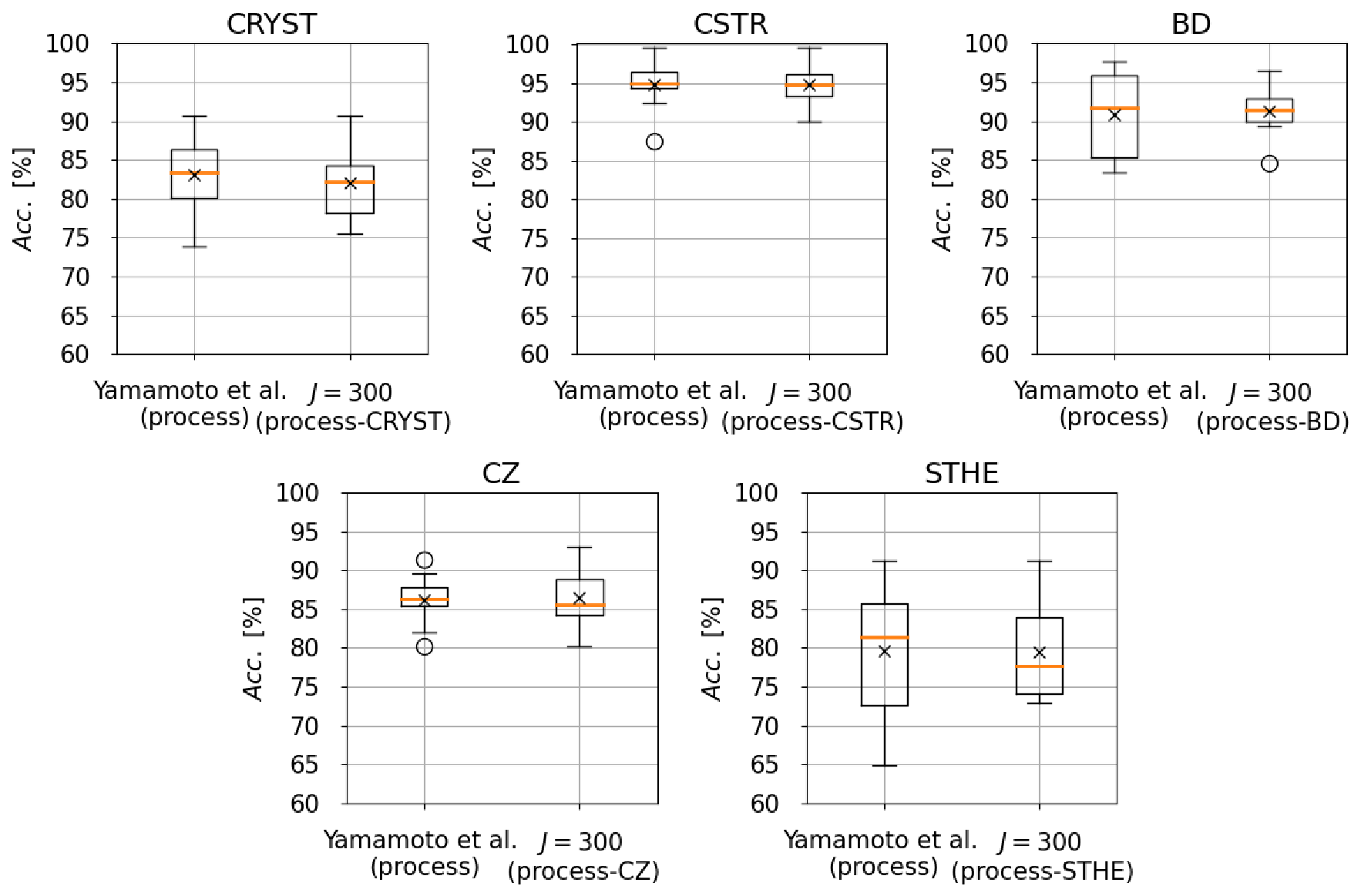}
  \caption{Distribution of $Acc.$ across ten experiments using two methods: the method by Yamamoto et al. with $\mathcal{D}_\mathrm{Process}$ as training data  (left side of each figure) and Proposed$_{J=300}$ with $\mathcal{D}_{\mathrm{Process}-X}$ as training data (right side of each figure).
  The orange line indicates the median and the mark $\times$ represents the mean.}
  \label{fig:pro-hige}
\end{figure}

Fig.~\ref{fig:pro-hige} shows the distribution of $Acc.$ across ten experiments for the method of Yamamoto et al. using $\mathcal{D}_\mathrm{Process}$ as training data, and Proposed$_{J=300}$ using $\mathcal{D}_{\mathrm{Process}-X}$ as training data.
The proposed method ($J=300$) using $\mathcal{D}_{\mathrm{Process}-X}$ achieved comparable $Acc.$ to the method of Yamamoto et al. using $\mathcal{D}_\mathrm{Process}$.
This suggests that applying the proposed method to $\mathcal{D}_\mathrm{Process}$ could achieve performance equivalent to the method of Yamamoto et al. without preparing training data for the target process.

\begin{figure}[t]
  \centering
  \includegraphics[width=.5\columnwidth]{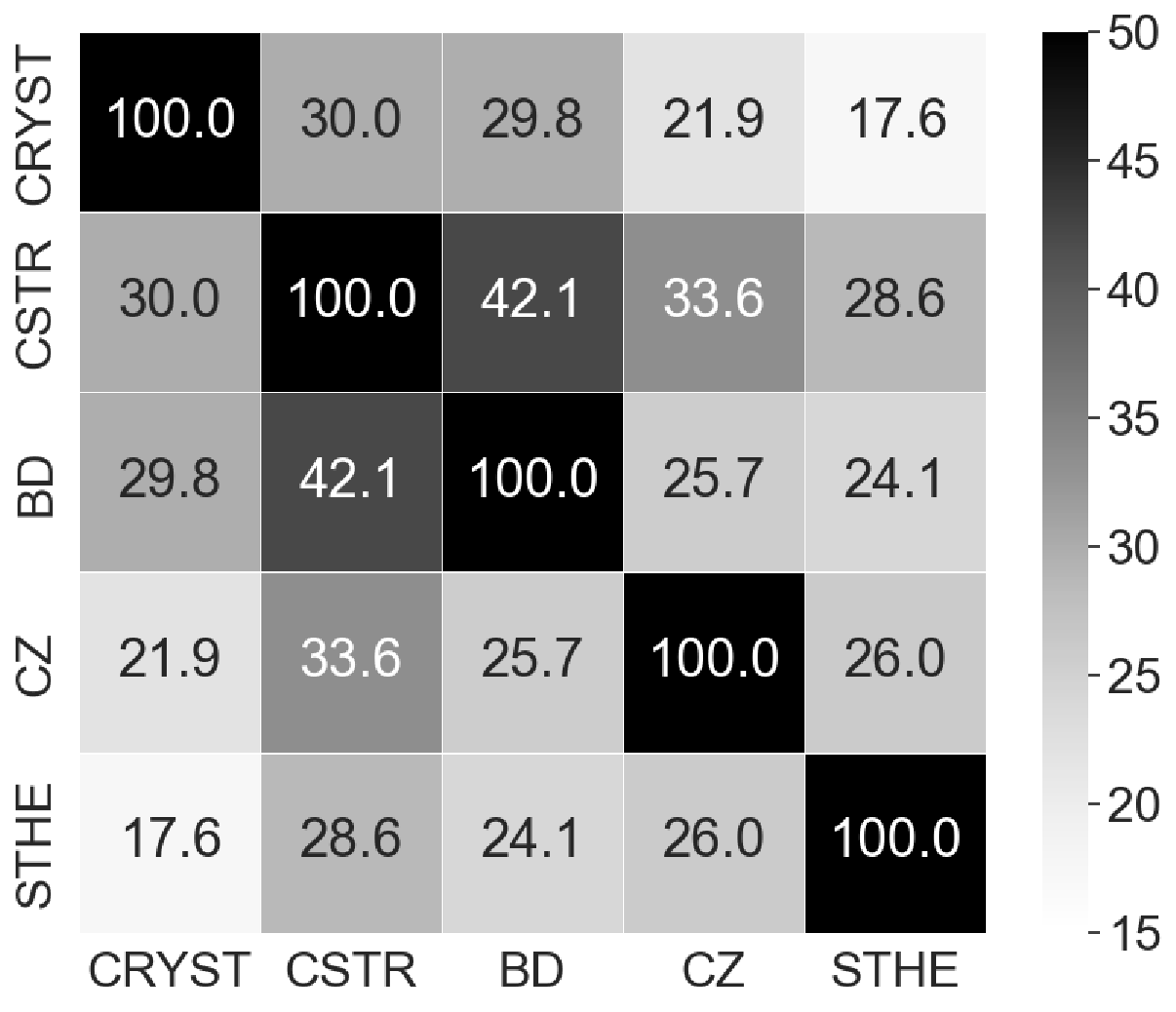}
  \caption{Similarities between datasets. Each value is displayed in \%}.
  \label{fig:similarity}
\end{figure}

Moreover, when $D_{\mathrm{CRYST}}$, $D_{\mathrm{CZ}}$, and $D_{\mathrm{STHE}}$ were used for the test set, Proposed~$_{J=300}$ showed a significant increase in $Acc.$ by using data of the target process for training, whereas such performance increase was small when $D_{\mathrm{CSTR}}$ and $D_{\mathrm{BD}}$ were used for the test set.

This could be due to the presence of datasets similar to $D_{\mathrm{CSTR}}$ and $D_{\mathrm{BD}}$ in $\mathcal{D}_{\mathrm{Process}-X}$, limiting the effect of preparing their training data.
To verify this hypothesis, we introduce the Simpson coefficient as the similarity between datasets $\mathcal{D}_{A}$ and $\mathcal{D}_{B}$ as follows:
 \begin{equation}
    S(\mathcal{D}_A, \mathcal{D}_B) = \frac{|\mathcal{W}_A \cap \mathcal{W}_B|}{\min\{|\mathcal{W}_A|, |\mathcal{W}_B|\}} , \label{eq:simpson}
\end{equation}
  where $\mathcal{W}_X$ ($X \in \{A, B \} $) is the set of words that compose the definitions in dataset $\mathcal{D}_X$, $|\mathcal{W}_X|$ is the number of elements in $\mathcal{W}_X$, and $\mathcal{W}_A \cap \mathcal{W}_B$ is the set of elements belonging to both $\mathcal{W}_A$ and $\mathcal{W}_B$.
  $\mathcal{W}_X$ does not include stop words like ``of'' and ``and''.
  For instance, when $\mathcal{D}_A$ consists of ``velocity of air'' and ``temperature of reactor'', $\mathcal{W}_A$ = \{``velocity'', ``air'', ``temperature'', ``reactor''\}.
The value of $S(\mathcal{D}_A, \mathcal{D}_B)$ closer to 1 means that $\mathcal{D}_A$ and $\mathcal{D}_B$ are more similar.
Fig.~\ref{fig:similarity} illustrates the similarities between datasets included in $\mathcal{D}_\mathrm{Process}$.
$\mathcal{D}_{\mathrm{CSTR}}$ and $\mathcal{D}_{\mathrm{BD}}$ had the highest similarity at 42.1\%, indicating a substantial overlap in the words that comprise their definitions. This overlap could impede performance improvements that might otherwise be achieved by preparing training data for these processes.
Conversely, the dataset $\mathcal{D}_{\mathrm{STHE}}$ had the largest performance gain by using training data with Proposed$_{J=300}$, but exhibited lower similarity compared to the other datasets in $\mathcal{D}_\mathrm{Process}$.
These findings suggest that the performance of variable definition extraction can be enhanced by preparing training data specific to the test data's process, particularly when there exists a low similarity between the process of the test data and those of the training data.

\subsubsection{Analysis of Definition Extraction Failure Samples}
Table~\ref{tab:error} shows the breakdown of definition extraction failure samples in ten experiments conducted using Proposed$_{J=300}$.
For FP$^{(1)}$ failures, we further categorize them into three classes: FP$^{(1)}_{\text{wide}}$, FP$^{(1)}_{\text{narrow}}$, and FP$^{(1)}_{\text{other}}$.
When the predicted definition range entirely encompasses the correct definition range, we classify it FP$^{(1)}_{\text{wide}}$ and when the predicted definition range is entirely contained within the correct definition range, we classify it as FP$^{(1)}_{\text{narrow}}$. For other cases, we classify it as FP$^{(1)}_{\text{other}}$.
Across all ten experiments, the total number of failures by Proposed$_{J=300}$ was equal to or less than that by the method of Yamamoto et al.

\begin{table}[t]
    \caption{Breakdown of definition extraction failure samples by error class in ten experiments using Proposed$_{J=300}$. Each value in parentheses indicates the difference in the number of failures compared to the method of Yamamoto et al.~\cite{yamamoto}.}
    \label{tab:error}
    \centering
    \small
    \begin{tabular}{lcccccccccc}
    \toprule & \multicolumn{10}{c}{Experiment number} \\
    \cmidrule(r){2-11}
    Error class & 1 & 2 & 3 & 4 & 5 & 6 & 7 & 8 & 9 & 10 \\
    \midrule
    \multirow{2}{*}{$FN$} & 34 & 19 & 38 & 22 & 13 & 19 & 12 & 7 & 20 & 24 \\
    & (+25) & (-2) & (+16) & (+9) & (-15) & (-5) & (-2) & (-10) & (-6) & (-3)  \\
    \midrule
    \multirow{2}{*}{FP$^{(1)}_{\text{wide}}$}
    & 4 & 7 & 4 & 4 & 2 & 3 & 1 & 3 & 3 & 3 \\
    & (-2) & (-2) & (-5) & (+2) & (-1) & (-5) & (-2) & ($\pm$0) & (+1) & (-3)  \\
    \midrule
    \multirow{2}{*}{FP$^{(1)}_{\text{narrow}}$} & 6 & 3 & 5 & 1 & 2 & 4 & 2 & 4 & 6 & 3 \\
    & (-4) & (-1) & ($\pm$0) & (-5) & (-1) & ($\pm$0) & (-1) & (+1) & (+1) & (-2)  \\
    \midrule
    \multirow{2}{*}{FP$^{(1)}_{\text{other}}$}  & 1 & 1 & 1 & 2 & 4 & 1 & 1 & 1 & 1 & 1 \\
    & ($\pm$0) & (-3) & (-2) & (+1) & (+3) & (-1) & (+1) & ($\pm$0) & ($\pm$0) & ($\pm$0)  \\
    \midrule
    \multirow{2}{*}{$FP^{(2)}$} & 9 & 12 & 11 & 2 & 17 & 15 & 13 & 18 & 14 & 7 \\
    & (-30) & (-5) & (-21) & (-21) & (-7) & (-13) & (-2) & (-2) & (+4) & (+1) \\
    \midrule
    \multirow{2}{*}{Total}  & 54 & 42 & 59 & 32 & 38 & 42 & 29 & 33 & 44 & 38 \\
    & (-11) & (-13) & (-12) & (-14) & (-21) & (-24) & (-6) & (-11) & ($\pm$0) & (-7)  \\
    \bottomrule
    \end{tabular}
\end{table}

To discuss the effectiveness and limitations of the proposed method, we analyzed three cases: 1) the case where the definition was correctly extracted only by Proposed$_{J=300}$, 2) the case where the definition was incorrectly extracted by Proposed$_{J=300}$ but correctly extracted by the existing method, and 3) the case where the definition was incorrectly extracted by both methods.
In what follows, the source sentences for extraction and definition extraction results are showcased.
In the following case, the target variable symbol is dashed, the correct definition (or [CLS] token if there is no definition) is underlined, the extraction results by Proposed$_{J=300}$ and the existing method are shown in bold and italic, respectively.

\begin{itemize}
    \item[$\blacksquare$] \textbf{Type 1}: \targetdef{\defproposed{[CLS]}} The \defexisting{temperature} is then lowered to \targetvar{${T_{\mathrm{fin}}}$} within the time duration $t_\mathrm{cool}$.
\end{itemize}
Type 1 represents cases where the method of Yamamoto et al. extracted an incorrect definition for variables without definitions in the text, while Proposed$_{J=300}$ made improvements by not extracting anything. Template sentences include ones like ``The rate of reaction is described by the equation: d [VAR\_1]/d$t$ = $k$*[$A$][$B$].'', which contain the variable token but not the corresponding definition token.
The enhanced ability to correctly identify variables without definitions in the text using the proposed method is attributed to the increase in variables without definitions in the training data.

\begin{itemize}
  \item[$\blacksquare$] \textbf{Type 2}: \defproposed{[CLS]} The \defexisting{\targetdef{reaction rate constants for the $i^{th}$ reaction}} are given by \targetvar{$k_i$}, $i=1,\ldots,6$,  modeled using the Arrhenius rate equations, whose values can be obtained from the parameters reported in Table 1.
\end{itemize}

Type 2 illustrates the most common cases where the method of Yamamoto et al. could extract the correct definition, but Proposed$_{J=300}$ did not extract anything. The presence of template sentences without definition tokens in the analysis of Type 1 might contribute to an increase in such failure cases. Creating template sentences with a lower ratio of variables without definitions might increase the number of variables judged to have definitions.

\begin{itemize}
  \item[$\blacksquare$] \textbf{Type 3}: \defexisting{\defproposed{[CLS]}} The total moments, seed moments, and \targetdef{nucleated crystal moments} are related by $\mu_{i,T}$ = $\mu_{i,S}$ + \targetvar{${\mu_{i,N}}$}, and the moments are defined in terms of their respective CSD's as
    \begin{align}
      \mu_{i,j}(t) = \int_0^\infty x^i n_j(x,t) \, dx, \quad i = 0, 1, 2, \ldots, j = T, S, N \quad (3) \nonumber
    \end{align}
  and have units of m$^i$/kg solvent.
\end{itemize}
Cases where both methods were incorrect included various patterns, but many involved sentences with a mix of multiple variables, definitions, and equations, where a variable with a definition was not extracted. Although the template sentences created in this study contain multiple variable and definition tokens, all have explicit variable-definition relationships and do not include equations. Developing methods of creating template sentences closer to the definition sentences found in papers on chemical processes could lead to further performance improvements.

\section{Conclusion}
In this study, we enhanced the performance of variable definition extraction by preparing template sentences for definitions and expanding the training data by assigning variable-definition pairs from the training data to each sentence.
Comparing the definition extraction performance of the existing model with that of the model trained by the proposed method, we found that increasing the number of template sentences improved the accuracy rate, achieving an accuracy of 88.3\%, which surpasses the existing method by 2.5 percentage points.
Moreover, by applying the proposed method, we achieved an accuracy rate equivalent to the existing method without using data from the definition extraction target process in training.

Future challenges are as follows. There were instances where the existing method could extract the correct definition, but the application of the proposed method resulted in no extraction. This could be due to an inappropriate ratio of template sentences that do not include definition tokens. Therefore, adjusting this ratio could lead to further performance improvements. Additionally, failure cases involving sentences with a mix of multiple variables, definitions, and equations were not improved by using the proposed method. Although the template sentences created in this study contain multiple variable and definition tokens, all have explicit variable-definition relationships and do not include equations. Generating template sentences more similar to the definition sentences appearing in the dataset could expect further performance improvements.


%
%
%

\bibliographystyle{splncs04}
\bibliography{reference}

\end{document}